\pdfoutput=1
\documentclass[11pt]{article}

\usepackage[]{ACL2023}

\usepackage{times}
\usepackage{latexsym}

\usepackage[T1]{fontenc}
\usepackage[utf8]{inputenc}

\usepackage{microtype}

\usepackage{inconsolata}

\usepackage{amsthm}
\usepackage{amsmath}
\usepackage{amssymb}
\usepackage{caption}
\usepackage{subcaption}
\usepackage{enumerate}
\usepackage{multirow}
\usepackage{graphicx} 
\usepackage{url}
\usepackage{booktabs}
\usepackage{color}
\usepackage{algorithm}
\usepackage{algorithmic}
\usepackage{makecell}

\usepackage{cleveref}
\crefname{section}{§}{§§}
\Crefname{section}{§}{§§}

\title{Query Structure Modeling for Inductive Logical Reasoning \\Over Knowledge Graphs}

\author{Siyuan Wang\textsuperscript{\rm 1}, Zhongyu Wei\textsuperscript{\rm 1,2}\thanks{~~Corresponding author}, \bf Meng Han\textsuperscript{\rm 3}, \\
\bf Zhihao Fan\textsuperscript{\rm 1}, Haijun Shan\textsuperscript{\rm 4}, Qi Zhang\textsuperscript{\rm 5}, Xuanjing Huang\textsuperscript{\rm 5} \\
\textsuperscript{\rm 1}School of Data Science, Fudan University, China \\
\textsuperscript{\rm 2}Research Institute of Intelligent and Complex Systems, Fudan University, China \\
\textsuperscript{\rm 3}Huawei Poisson Lab, China \textsuperscript{\rm 4}CEC GienTech Technology Co., Ltd, China \\\textsuperscript{\rm 5}School of Computer Science, Fudan University, China \\
\{wangsy18,zywei,fanzh18,qz,xjhuang\}@fudan.edu.cn
\\hanmeng12@huawei.com; haijun.shan@gientech.com \\
}

\begin{document}
\maketitle
\begin{abstract}
Logical reasoning over incomplete knowledge graphs to answer complex logical queries is a challenging task. With the emergence of new entities and relations in constantly evolving KGs, inductive logical reasoning over KGs has become a crucial problem. 
However, previous PLMs-based methods struggle to model the logical structures of complex queries, which limits their ability to generalize within the same structure.
% Recent embedding-based methods that map query structures and entities into a vector space have shown remarkable performance in logical reasoning over KGs. However, they only focus on the transductive setting and are unable to generalize to new emerging entities or relations. 
In this paper, we propose a structure-modeled textual encoding framework for inductive logical reasoning over KGs. It encodes linearized query structures and entities using pre-trained language models to find answers. For structure modeling of complex queries, we design stepwise instructions that implicitly prompt PLMs on the execution order of geometric operations in each query. We further separately model different geometric operations (i.e., projection, intersection, and union) on the representation space using a pre-trained encoder with additional attention and maxout layers to enhance structured modeling.
We conduct experiments on two inductive logical reasoning datasets and three transductive datasets. The results demonstrate the effectiveness of our method on logical reasoning over KGs in both inductive and transductive settings. \footnote{Codes are publicly available at \url{https://github.com/WangsyGit/InductiveLR}.} 
\end{abstract}

\section{Introduction}
\begin{figure}[!th]
\centering
\includegraphics[width=1.0\columnwidth]{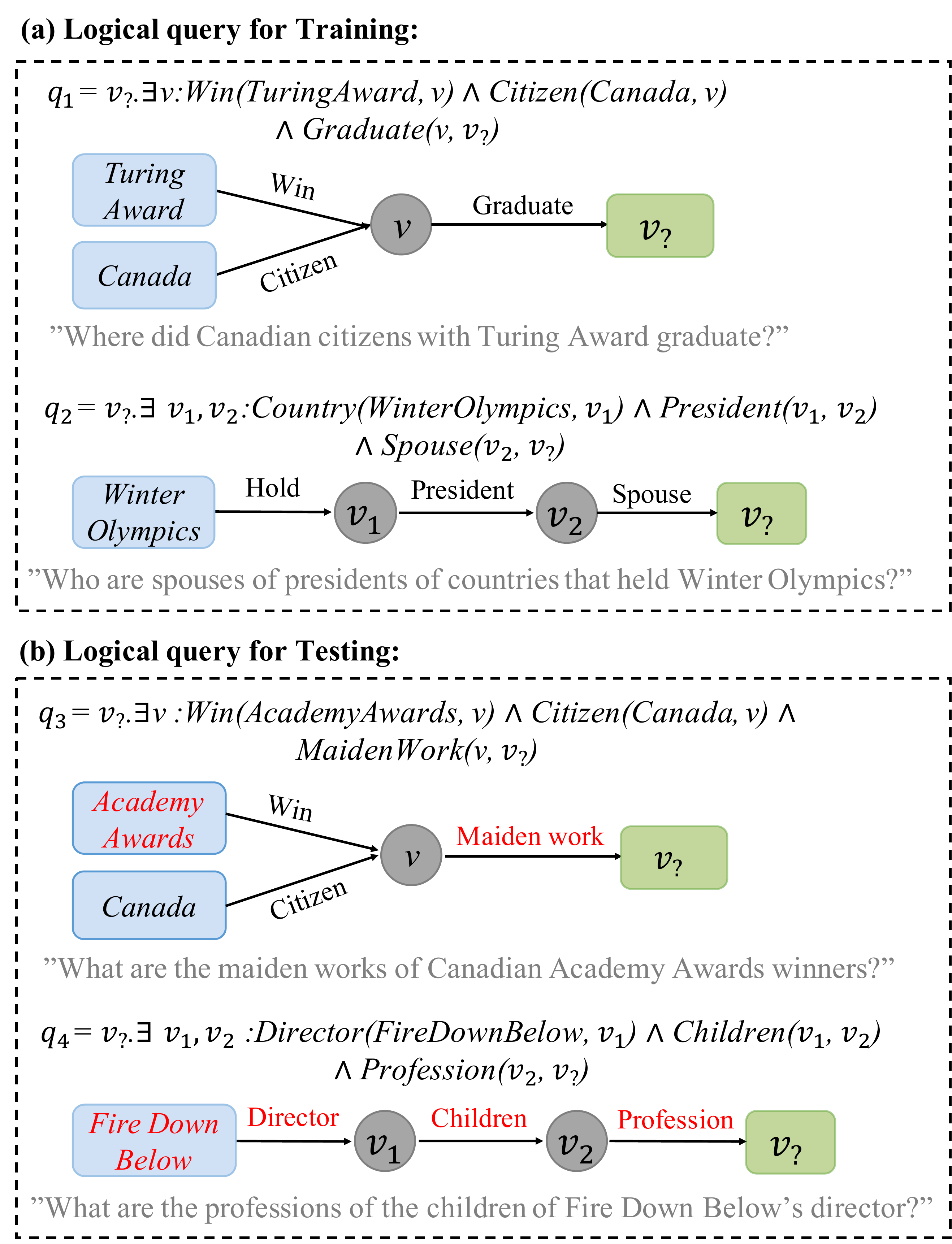}
\caption{\label{figure1} Examples of inductive logical reasoning over KGs: testing queries contain unseen entities and relations (in red) during training. Each query is associated with an intrinsic logical structure and its natural language interpretation.
}
\end{figure}

%Recent years have witnessed increasing attentions on logical reasoning over widely used KGs such as Freebase~\cite{bollacker2008freebase}, Yago~\cite{suchanek2007yago}, NELL~\cite{carlson2010toward} and Wikidata~\cite{vrandevcic2014wikidata}. 

Logical reasoning over knowledge graphs (KGs) aims to answer complex logical queries given large-scale KGs~\cite{guu2015traversing, hamilton2018embedding}.
Recent years have witnessed increasing attention on logical reasoning over widely used KGs such as Freebase~\cite{bollacker2008freebase}, Yago~\cite{suchanek2007yago}, NELL~\cite{carlson2010toward} and Wikidata~\cite{vrandevcic2014wikidata}. 
% For example, the missing relation (Hinton, graduate, Edinburgh) will make it difficult to answer the logical query in Figure~\ref{figure1}(a) 
% ``$q=v_?.\exists v:Win (TuringAward, v)\wedge Citizen (Canada, v)\wedge Graduate (v, v_?)$'' 
% (i.e., ``Where did Canadian citizens with Turing Award graduate?''). 
With missing relations in the KG, it is challenging to deduce correct answers to complex queries by traversing the graph. Previous work primarily focuses on transductive logical reasoning where the training and testing are done on the same KG with the same group of entities. They typically rely on geometric embedding-based methods to map both entities and queries into a joint low-dimensional vector space~\cite{hamilton2018embedding,ren2020query2box, ren2020beta}. The goal is to push the embeddings of answer entities and queries to be close to each other, allowing answers to be predicted through embedding similarity even when the involved relation is absent. 
% For logical reasoning, these methods model structural knowledge of complex queries by iteratively conducting geometric operations over embeddings.
In contrast, the inductive setting of logical reasoning has been rarely studied which requires generalizing to unseen entities and relations or even new KGs.
As real-world KGs are usually dynamic with emerging unseen entities and relations, it's significant to explore the inductive setting for complex query answering. 

\begin{figure*}[!th]
\centering
\includegraphics[width=2.05\columnwidth]{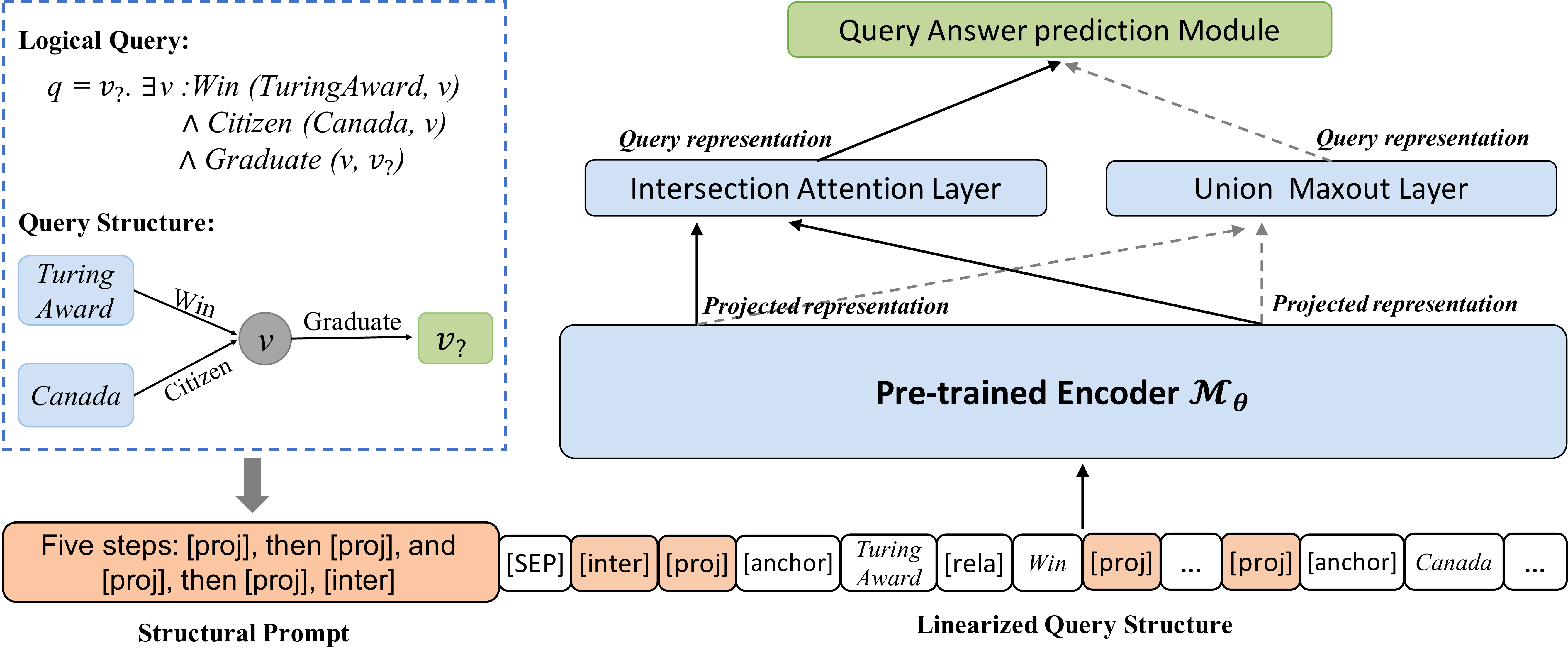}
\caption{\label{figure_framework} The overall architecture of our method for inductive KG logical reasoning with structure knowledge modeling. [inter] and [proj] are respectively abbreviations of [intersection] and [projection].}
\end{figure*}

Existing research on inductive logical reasoning mainly follows two directions. The first inherits embedding-based methods and incorporates type as additional information to improve inductive capability~\cite{hu2022type}, which can not generalize to unseen types of entities and relations.
%~\cite{yao2019kg, zha2022inductive} and explore link prediction tasks of one-step reasoning and typically . This results in incompetence in complex structure modeling for inductive logical reasoning involving complex queries. 
The second direction leverages pre-trained language models (PLMs) to encode textual information of entities/relations for generalization to unseen elements~\cite{wang2021kepler, daza2021inductive, wang2021structure}. PLMs-based approaches provide more flexible solutions and generate better results. However, they only explore link prediction tasks of one-step reasoning, and simply linearize the triplet or subgraph into text sequence without modeling explicit reasoning structure~\cite{yao2019kg, zha2022inductive}.
%~\cite{yao2019kg, zha2022inductive} and explore link prediction tasks of one-step reasoning and typically . This results in incompetence in complex structure modeling for inductive logical reasoning involving complex queries. 
% For inductive logical reasoning involving complex queries, structure modeling of different query types is crucial. An inductive example is shown in Figure~\ref{figure1}. The query $q_1$ and $q_2$ which are both composed of three triples appear to have a similar format but actually have different logical structures. We argue that modeling the intrinsic logical structures of complex queries can help logical reasoning generalization within the same structure.
% For inductive logical reasoning involving complex queries, structure modeling of different query types is crucial. 
An example is shown in Figure~\ref{figure1}. Two findings stand out. (1) The query $q_1$ and $q_2$ appear to be similar in format (both as a conjunction of three terms) but actually have different logical structures. PLMs-based methods that encode flattened queries can not model this structure information for correct logical reasoning. (2) Although queries $q_1$ and $q_3$ (also $q_2$ and $q_4$) contain different elements, they share the same logical structure.
% Although the testing query $q_1(b)$ and the training query $q_2(a)$ share several elements, the logical reasoning method from $q_2(a)$ can not be directly transferred to $q_1(b)$. 
Motivated by these, we argue that structure modeling of different complex queries can further boost the generalization ability of logical reasoners.

In this paper, we propose to model query structure for inductive logical reasoning over KGs. Specifically, we transform the query structure into a sequence using textual names of involved entities, relations, and logical operators. For complex query structures composed of multiple geometric operations over entities and relations, we introduce two measures to enable logical structure modeling during text encoding.
First, we design stepwise instructions for different query types to indicate which operation in the query structure should be conducted at each step and feed them as the structural prompt to PLMs. 
Besides, we extend the pre-trained encoder with an additional attention layer and a maxout layer to respectively model different geometric operations including projection, intersection, and union on the representation space, to implicitly inject structured modeling into PLMs. Our proposed method is a generic inductive framework, which can be plugged into different PLMs for better performance.

% with the structural knowledge of complex queries preserved. 

%For XXXXXXX, we follow the textual encoding paradigm with the force of PLMs to encode the linearized query and candidate entities, and make their representations as similar as possible. Specifically, we transform the query structure into a sequence using textual names of involved entities, relations, and logical operators. 
We conduct experiments on two datasets for inductive logical reasoning over KGs, FB15k-237-V2 and NELL-V3~\cite{teru2020inductive} as well as three transductive datasets, FB15k~\cite{bordes2013translating}, FB15k-237~\cite{toutanova2015observed}, and NELL995~\cite{xiong2017deeppath}.  
The results demonstrate that our method achieves strong inductive performance on unseen entities and relations, even across different KGs, without sacrificing logical reasoning capability and generalizability to new query structures.    

\section{Methodology}
In this work, we study the task of complex logical reasoning over KGs. The input is a first-order logic query $q$ which can include any set of existential quantification ($\exists$), conjunction ($\wedge$), and disjunction ($\vee$) operators (such as the query in Figure~\ref{figure1}). Our goal is to predict a set of entities $\mathcal{A}=\{a_1,a_2,a_3...\}$ that answer the query $q$ based on an incomplete KG $\mathcal{G=(E, R)}$ which consists of a set of triplets $(h, r, t)$ but lacks several involved relations. Here $h,t \in \mathcal{E}$ are the head and tail entities and $r \in \mathcal{R}$ is the relation between them.
We mainly focus on the inductive setting of KG logical reasoning, where the evaluated queries contain entities/relations that are completely unseen during the training period. 

%As illustrated in Figure~\ref{figure_framework}, we propose to encode the text sequences of query structures and predict the answer entities through representation similarity for inductive logical reasoning over KGs. 

%To preserve the structure knowledge of complex queries during textual encoding, we introduce stepwise instructions to prompt PLMs to implicitly conduct different geometric operations sequentially. Additionally, we adopt the pre-trained encoder, an attention module and a maxout layer to respectively model three geometric operations (i.e., projection, intersection and union) for better structural learning.

Figure~\ref{figure_framework} shows the overall architecture of our model. We propose to encode the text sequences of query structures and predict the answer entities based on representation similarity for inductive logical reasoning over KGs. In this section, we first list different types of query structures studied in logical reasoning over KGs(\cref{sec:query_structure_type}). Then according to various query types, we introduce our structure linearization and structural prompts for textual encoding(\cref{sec:query_structure_encode}). The geometric operation modeling and query answer prediction modules are described in (\cref{sec:geometric_operation}) and (\cref{sec:answer_prediction}). Finally, we provide the details about training and inference (\cref{sec:training_inference}). 

\subsection{Query Structure Types}
\label{sec:query_structure_type}
Following~\cite{ren2020query2box}, we consider 9 types of complex query structures which are composed of different sets of geometric operations (including projection, intersection and union) over entities and relations. These include six single-operation query structures and three mixed-operation ones. Specifically, three query types only focus on projection, including one-relation projection (1p), two-relation projection (2p), and three-relation projection (3p). Two query types focus on the intersection of two triplets (2i) and three triplets (3i), and another one focuses on the union of two triplets (2u). The three mixed-operation query structures are respectively the combinations of intersection\&projection (ip), projection\&intersection (pi), and union\&projection (up). The different query structures are illustrated as the following formula:
\begin{align}
    [1p] \ q=v_?&:r_1(e_1,v_?) \nonumber \\
    [2p] \ q=v_?&.\exists v:r_1(e_1,v)\wedge r_2(v,v_?) \nonumber \\ 
    [3p] \ q=v_?&.\exists v_1,v_2:r_1(e_1,v_1)\wedge r_2(v_1,v_2) \nonumber \\ 
    &\wedge r_3(v_2,v_?) \nonumber \\ 
    [2i] \ q=v_?&:r_1(e_1,v_?)\wedge r_2(e_2,v_?) \nonumber \\ 
    [3i] \ q=v_?&:r_1(e_1,v_?)\wedge r_2(e_2,v_?)\wedge r_3(e_3,v_?) \nonumber \\ 
    [pi] \ q=v_?&.\exists v:r_1(e_1,v)\wedge r_2(v,v_?)\wedge r_3(e_2,v_?) \nonumber \\
    [ip] \ q=v_?&.\exists v:r_1(e_1,v)\wedge r_2(e_2,v)\wedge r_3(v,v_?) \nonumber \\ 
    [2u] \ q=v_?&:r_1(e_1,v_?)\vee r_2(e_2,v_?) \nonumber \\ 
    [up] \ q=v_?&.\exists v:(r_1(e_1,v)\vee r_2(e_2,v))\wedge r_3(v,v_?) 
\end{align}
where $e_i$ and $v_i$ are the anchor entities and existentially quantified bound variables entities, and $v_?$ are the target answer entities to the query.
As these complex queries contain rich logical structural information, we need to model the structure knowledge during textual encoding for better inductive generalization within the same logical structure.   

\subsection{Query Structure Encoding}
\label{sec:query_structure_encode}
In order to use PLMs for better generalization to unseen entities/relations, we first need to linearize the query structures into text sequences. We also design instructions for each query type as a structural prompt to implicitly indicate the order of geometric operations execution to PLMs.
We concatenate the linearized query and structural prompt as the input, and encode them to obtain the query representation for matching with the answer entities.

\paragraph{Query Structure Linearization}
Given a query structure $q$, we customize its linearization method according to the query type. For each triplet $r_i(e_i, v)$ in the query, we formulate it as ``\texttt{[anchor] $t(e_i)$ [relation] $t(r_i)$}'' and 
ignores the intermediate variable entities, where $t(e_i)$ and $t(r_i)$ are the textual names of anchor entity $e_i$ and relation $r_i$. Then we add the names of logical operations before the involved subqueries. For example, the query structure of type [2p] can be linearized into sequence ``\texttt{[projection] [anchor] $t(e_1)$ [relation] $t(r_1)$ [projection] [relation] $t(r_2)$}'' and the query structure of type [2i] can be mapped into ``\texttt{[intersection] [projection] [anchor] $t(e_1)$ [relation] $t(r_1)$ [projection] [anchor] $t(e_2)$ [relation] $t(r_2)$}''.

For query types [ip] and [up] that are composed of intersection/union and projection, the last relation projection is conducted over the intersection/union of previous triplets. Directly flattening the query is unable to keep such structural information. Therefore, we propose to split the intersection/union and repeatedly connect each triplet with the last relation projection which moves the intersection/union operation to the final step. This transformation is equivalent to the original query structure. For example, the following two query structures are equivalent and are both of type [up]. 
\begin{align}
    \label{union_transformation}
    &(r_1(e_1,v)\vee r_2(e_2,v))\wedge r_3(v,v_?) \nonumber \\
    &(r_1(e_1,v)\wedge r_3(v,v_?))\vee (r_2(e_2,v)\wedge r_3(v,v_?))
\end{align}
Based on this transformation, we linearize the query structure of type [up] into the text sequence as ``\texttt{[union] [projection] [anchor] $t(e_1)$ [relation] $t(r_1)$ [projection] [relation] $t(r_3)$ [projection] [anchor] $t(e_2)$ [relation] $t(r_2)$ [projection] [relation] $t(r_3)$}''. The details of structure linearization templates for each query type are listed in Appendix~\ref{templates}.

\paragraph{Structural Prompt} Besides feeding the linearized query structure into PLMs, we also introduce stepwise instructions that indicate the order of geometric operations execution to prompt the pre-trained encoder with implicit structural information of the query. Specifically, each prompt consists of two parts: the number of total execution steps and the operation order in the query, which is formulated as ``\texttt{total steps: operation order}''. For query types [ip] and [up], we define the total steps and operation order according to the transformed query structure in Eq~\ref{union_transformation}. 
% We list these two parts of each query type in Table~\ref{prompt_details}. 
The detailed structural prompts of each query type are presented in Table~\ref{table_prompts}.
% to instruct the pre-trained language model to implicitly execute involved logical operations stepwise. 
% \begin{table}[!h]
%     \centering
%     \begin{tabular}{|c|c|c|}
%     \hline
%       Query Type & Total Steps & Operation Order \\
%     \hline
%         \text{[1p]} & 1 & p \\
%         \text{[2p]} & 2 & p,p \\
%         \text{[3p]} & 3 & p,p,p \\
%         \text{[2i]} & 3 & p,p,i \\
%         \text{[3i]} & 4 & p,p,p,i \\
%         \text{[pi]} & 4 & p,p,p,i\\
%         \text{[ip]} & 5 & p,p,p,p,i\\
%         \text{[2u]} & 3 & p,p,u\\
%         \text{[up]} & 5 & p,p,p,p,u\\
%     \hline
%     \end{tabular}
%     \caption{\label{prompt_details} Total steps and operation execution orders of different query types. ``p'', ``i'' and ``u'' are the abbreviations of ``projection'', ``intersection'' and ``union''.}
%     \label{tab:my_label}
% \end{table}

\begin{table}[!h]
    \begin{center}
    \resizebox{0.48\textwidth}{!}{
    \begin{tabular}{m{1.9cm}m{7.5cm}}
    \toprule 
    \bf Query Type & \bf Structural Prompt \\
    \midrule
    \quad \ \text{[1p]} & One step: [proj] \\
    \midrule
    \quad \ \text{[2p]} & Two steps: [proj], then [proj] \\
    \midrule
    \quad \ \text{[3p]} & Three steps: [proj], then [proj], then [proj] \\
    \midrule
    \quad \ \text{[2i]} & Three steps: [proj], and [proj], [inter] \\
    \midrule
    \quad \ \text{[3i]} & Four steps: [proj], and [proj], and [proj], [inter] \\
    \midrule
    \quad \ \text{[pi]} & Four steps: [proj], then [proj], and [proj], [inter] \\
    \midrule
    \quad \ \text{[ip]} & Five steps: [proj], then [proj], and [proj], then [proj], [inter] \\
    \midrule
    \quad \ \text{[2u]} & Three steps: [proj], and [proj], [union]  \\
    \midrule
    \quad \ \text{[up]} & Five steps: [proj], then [proj], and [proj], then [proj], [union] \\
    \bottomrule
    \end{tabular}
    }
\caption{\label{table_prompts} Structural prompts of sifferent query types. ``[proj]'' and ``[inter]'' are the abbreviations of ``[projection]'' and ``[intersection]''.}
    \end{center}
\end{table}

% Based on different query types, we obtain corresponding structural prompts. For example, the structural prompt of query type [up] is ``\texttt{Five steps: [projection], then [projection], and [projection], then [projection], [union]}''. Other prompts can be found in Appendix~\ref{structural_prompt}.
Then the input $t$ can be obtained by concatenating the structural prompt $s$ and the linearized query structure $t(q)$, which is formulated as ``\texttt{[CLS] [qtype] s [SEP] t(q)}''. We feed it to the pre-trained encoder and obtain the output hidden states $H=(h_1,h_2, ..., h_{|t|})$.
% take the average pooling of output hidden states as the query representation $h_q$. 

\subsection{Geometric Operation Modeling}
\label{sec:geometric_operation}
To further enhance the structure modeling during textual encoding, we propose to separately model different geometric operations in logical queries to explore their spatial characteristics. 
As Transformer-based encoders are widely used to implicitly learn the translation function for simple link prediction and question answering~\cite{yao2019kg, wang2021structure}, we directly utilize it for modeling multi-hop relation projection. For each path of relation projection $r_1(e_1,v_1)\wedge...\wedge r_j(v_j,v_?)$, we extract the hidden states corresponding to sequence ``\texttt{[anchor] $t(e_1)$ [relation] $t(r_1)$ ... [projection] [relation] $t(r_j)$}'' from $H=(h_1,h_2, ..., h_{|t|})$. We then take the average as the representation of target entity $v_?$, which also can be viewed as the representation $h_q$ of the query that only involves relation projection operation.

For the intersection and union of multiple subqueries, we adopt an attention layer~\cite{bahdanau2014neural} and a maxout layer~\cite{goodfellow2013maxout} on top of the pre-trained encoder to respectively model these two operations in the geometric representation space. Specifically, we feed the representations of target entities in all subqueries to these two additional layers to achieve the intersection and union operations. The output can be taken as the query representation $h_q$ that contains intersection or union.

As presented in~\cite{ren2020beta}, the complete set of first-order logic operations encompasses existential quantification ($\exists$), conjunction ($\wedge$), disjunction ($\vee$) and negation($\neg$). Our approach covers the first three operations by modeling relation projection, intersection, and union respectively. The negation operation is not individually modeled as pre-trained encoders are capable of capturing semantic exclusion for negation. We can add negative terms such as ``not'' before the corresponding relations within the input and feed it into pre-trained encoders to naturally address this task.

\subsection{Query Answer Prediction}
\label{sec:answer_prediction}
To answer the complex query $q$, we adopt the Siamese dual encoder~\cite{gillick2018end} to respectively encode the query $q$ and the candidate entity $c_i$ to match the answer.
We formulate the entity input as the sequence ``\texttt{[CLS] [target] $t(c_i)$}'' and feed it into the pre-trained encoder to obtain the candidate entity representation $h_{c_i}$ by taking the average of the hidden states.
Then we compute the similarity $d_i$ between the query representation $h_q$ and entity representation $h_{c_i}$, and encourage the query representation to be similar to the positive answer entities while dissimilar to negative entities. The entities whose representations are similar enough to the query will be predicted as the answers. We can pre-compute the representations of all candidate entities and utilize them to predict answers for different queries more efficiently.

% Given all candidate entities $(c_1, c_2, ..., c_N)$, we apply a softmax function to the cosine similarities between them and the query to calculate the plausibility score $s^1_{c_i}$ of each entity $c_i$ to be correct answers. 

The above matching scheme can handle the inductive setting when the candidate entity set is not closed and new entities may arise.
To improve the answer prediction accuracy in the transductive setting where the candidate entity set is closed, we also employ a classification layer on top of the query representation $h_q$. Given the fixed candidate entity set $(c_1, c_2, ..., c_N)$, 
the classification layer with a softmax function will output an $N$-dimensional plausibility distribution $(s_{c_1}, s_{c_2}, ..., s_{c_N})$ for each candidate entity $c_i$ to be the answer.

\subsection{Training \& Inference}
\label{sec:training_inference}
We simultaneously optimize a matching objective and a classification objective to train our inductive model for answering complex logical queries. For the former, we adopt contrastive learning~\cite{chen2020simple} which needs to separate the positive and negative answers for the query. We take the given ground-truth answer $c_+$ as positive and implement in-batch negative sampling to collect the negatives. 
We measure the similarity between the query and entities using dot product, and follow~\cite{he2020momentum} to utilize InfoNCE as the contrastive loss. The loss function is formulated as Eq.~\ref{loss_1} where $\tau$ is
\begin{align}
    \mathcal{L}_{M} &= -\log \frac{\exp(h_q \cdot h_{c_+} / \tau)}
    {\sum_{i=1}^{N} (\exp(h_q \cdot h_{c_i} / \tau)}
    \label{loss_1}
\end{align}
the temperature hyper-parameter and $N$ is the total number of candidate entities including positives and negatives. 
For the classification objective, we take all entities in each KG as candidate entities and calculate the cross-entropy loss as  Eq.~\ref{loss_2}.
\begin{align}
    \mathcal{L}_{C} = -\log \frac{s_{c_+}}{\sum_{i=1}^N \exp(s_{c_i})}
    \label{loss_2}
\end{align}
These two losses are combined in a weighted manner as $\mathcal{L}=\mathcal{L}_{M}+\lambda \mathcal{L}_{C}$ and $\lambda$ is the weighted hyper-parameter.

% to be edited
During inference, we perform differently for inductive and transductive logical query answering. 
For the inductive reasoning, we utilize the matching scheme and rank the representation similarities between the query and all candidate entities for query answer prediction.
For the transductive inference, we only adopt the classification scheme and find the most
plausible answer according to classification scores.
% and matching scores of candidate entities to infer the answers. 
% To get the matching score, we also apply a softmax function to the similarities between all candidate entities and the query to calculate the plausibility score $s^m_{c_i}$ of each entity $c_i$. Then we take $\max\{s^m_{c_i}, s_{c_i}\}$ as the final score of entity $c_i$.

\section{Experiments}
\subsection{Experiment Setup}
We conduct experiments on complex logical reasoning over KGs in both inductive and transductive setting. 
For the inductive setting, we adopt two datasets, FB15k-237-V2 and NELL-V3, that have disjoint sets of entities for training and evaluation, as introduced by~\cite{teru2020inductive}. To further challenge our model, we also illustrate the cross-KG inductive generalization performance by respectively taking FB15k and NELL995 for training/inference and inference/training
that contain completely different entities and relations.
In the transductive setting, we evaluate our model on the generated queries from three datasets: FB15k~\cite{bordes2013translating}, FB15k-237~\cite{toutanova2015observed}, and NELL995~\cite{xiong2017deeppath}, as proposed by~\cite{ren2020query2box}. 
All these datasets cover nine types of query structures. We follow the setting of ~\cite{ren2020query2box} to illustrate the generalization within the same structure to unseen entities and relations, and also the generalization to more complicated unseen structures composed of different structures. Specifically, we train our model on the first five types of query structures (1p, 2p, 3p, 2i, 3i) and evaluate it on all nine query types (1p, 2p, 3p, 2i, 3i, pi, ip, 2u, up), including both seen and unseen query structures during training. 
\begin{table*}[th]
\begin{center}
\resizebox{0.94\textwidth}{!}{
\begin{tabular}{l|c|ccc|cc|cccc}
\toprule
Model & Avg & 1p & 2p & 3p & 2i & 3i & ip & pi & 2u & up \\
\midrule
\multicolumn{11}{c}{ FB15k-237-V2 } \\
\midrule
\emph{Q2B} & 0.043 & 0.005 & 0.055 & 0.017 & 0.007 & 0.007 & 0.078 & 0.129 & 0.027 & 0.061 \\
\emph{TEMP(GQE)} & 0.163 & 0.146 & 0.221 & 0.141 & 0.139 & 0.144 & 0.157 & 0.220 & 0.097 & 0.201 \\
% \emph{Ours} & \bf 0.309 & \bf 0.392 & 0.308 & 0.234 & \bf 0.316 & \bf 0.343 & \bf 0.328 & \bf 0.346 & \bf 0.229 & \bf 0.284 \\
\emph{BiQE} & 0.158 & 0.286 & 0.151 & 0.107 & 0.187 & 0.240 & 0.125 & 0.155 & 0.085 & 0.091 \\
\emph{SILR} & \bf 0.178 & \bf 0.309 & 0.121 & 0.106 & \bf 0.237 & \bf 0.274 & 0.148 & 0.181 & \bf 0.110 & 0.116 \\
\midrule
\multicolumn{11}{c}{ NELL-V3 } \\
\midrule
\emph{Q2B} & 0.017 & 0.002 & 0.022 & 0.005 & 0.003 & 0.002 & 0.026 & 0.048 & 0.018 & 0.028 \\
\emph{TEMP(GQE)} & 0.062 & 0.096 & 0.057 & 0.060 & 0.072 & 0.081 & 0.047 & 0.062 & 0.040 & 0.040 \\
\emph{BiQE} & 0.089 & 0.178 & 0.081 & 0.081 & 0.082 & 0.092 & 0.067 & 0.081 & 0.066 & 0.069 \\
% \emph{BiQE} & 0.079 & 0.181 & 0.069 & 0.067 & 0.081 & 0.067 & 0.067 & 0.069 & 0.055 & 0.058 \\
\emph{SILR} & \bf 0.101 & \bf 0.197 & 0.079 &  0.074 & \bf 0.103 & \bf 0.122 & \bf 0.094 & \bf 0.090 & \bf 0.080 &  0.068 \\
\bottomrule
\end{tabular}
}
\caption{\label{table_result_inductive1} Inductive H@10 results of different structured queries on FB15k-237-V2 and NELL-V3 datasets. The results of \emph{Q2B} and \emph{TEMP(Q2B)} are taken from~\cite{hu2022type}. Avg is the average performance of all query types.}
\end{center}
\end{table*}
\begin{table*}[th]
\begin{center}
\resizebox{0.92\textwidth}{!}{
\begin{tabular}{l|c|ccc|cc|cccc}
\toprule
Model & Avg & 1p & 2p & 3p & 2i & 3i & ip & pi & 2u & up \\
\midrule
\multicolumn{11}{c}{ FB15k $\rightarrow$ NELL995} \\
\midrule
% \emph{Q2B} & 0.043 & 0.005 & 0.055 & 0.017 & 0.007 & 0.007 & 0.078 & 0.129 & 0.027 & 0.061 \\
% \emph{TEMP(GQE)} & 0.163 & 0.146 & 0.221 & 0.141 & 0.139 & 0.144 & 0.157 & 0.220 & 0.097 & 0.201 \\
\emph{BiQE} & 0.042 & 0.077 & 0.076 & 0.054 & 0.025 & 0.023 & 0.036 & 0.035 & 0.016 & 0.039 \\
\emph{SILR} & \bf 0.069 & \bf 0.166 & \bf 0.101 & \bf 0.096 & \bf 0.033 & \bf 0.040 & 0.020 & 0.027 & \bf 0.063 & \bf 0.073 \\
\midrule
\multicolumn{11}{c}{ NELL995 $\rightarrow$ FB15k } \\
\midrule
% \emph{Q2B} & 0.017 & 0.002 & 0.022 & 0.005 & 0.003 & 0.002 & 0.026 & 0.048 & 0.018 & 0.028 \\
% \emph{TEMP(GQE)} & 0.062 & 0.096 & 0.057 & 0.060 & 0.072 & 0.081 & 0.047 & 0.062 & 0.040 & 0.040 \\
\emph{BiQE} & 0.082 & 0.175 & 0.103 & 0.078 & 0.082 & 0.094 & 0.041 & 0.057 & 0.033 & 0.072 \\
\emph{SILR} & \bf 0.098 & \bf 0.218 & \bf 0.128 & \bf 0.100 & 0.081 & 0.089 & 0.029 & \bf 0.059 &\bf 0.115 & 0.065 \\
\bottomrule
\end{tabular}
}
\caption{\label{table_result_inductive2} Inductive H@10 results on cross-KG generalization. FB15k $\rightarrow$ NELL995 and NELL995 $\rightarrow$ FB15k respectively mean that the model is trained on FB15k and tested on NELL995, and vice versa.}
\end{center}
\end{table*}
The data split statistics of logical
queries in these datasets are provided in Appendix~\ref{data_statistics}.

\subsection{Implementation Details}
We take bert-large-cased and bert-base-cased~\cite{devlin2018bert} as the pre-trained encoder for encoding the query structure in (FB15k-237-V2, NELL-V3) and (FB15k, FB15k-237, and NELL995), respectively.
All models are implemented using Huggingface~\cite{wolf2019huggingface},
% The batch size is 1084 and the negative sample size is 270. All models 
and trained for 30 epochs on 4 NVIDIA Tesla V100 GPUs with 16 GB of memory. The Adam is taken as the optimizer and the learning rate is 1.5e-4. We use a linear learning rate scheduler with $10\%$ warmup proportion. The weight hyper-parameter to balance losses is set to $\lambda = 0.3$ or $\lambda = 0.4$. For automatic evaluation, we use Hits@$K$ ($K=3,10$) as the metrics, which calculate the proportion of correct answer entities ranked among the top-$K$.

\begin{table*}[th]
\begin{center}
\resizebox{0.92\textwidth}{!}{
\begin{tabular}{l|c|ccc|cc|cccc}
\toprule
Model & Avg & 1p & 2p & 3p & 2i & 3i & ip & pi & 2u & up \\
\midrule
\multicolumn{11}{c}{ FB15k-237 } \\
\midrule
\emph{GQE} & 0.228 & 0.402 & 0.213 & 0.155 & 0.292 & 0.406 & 0.083 & 0.17 & 0.169 & 0.163 \\
\emph{Q2B} & 0.268 & 0.467 & 0.240 & 0.186 & 0.324 & 0.453 & 0.108 & 0.205 & 0.239 & 0.193 \\
\emph{TEMP(Q2B)} & 0.294 & 0.457 & 0.278 & 0.234 & 0.369 & 0.496 & 0.229 & 0.117 & 0.276 & 0.189 \\
\emph{BiQE} & 0.262 & 0.472 & 0.293 & 0.245 & 0.345 & 0.473 & 0.096 & 0.211 & 0.075 & 0.145 \\
% \emph{Ours} & \bf 0.273 & 0.459 & 0.266 & 0.217 & 0.342 & \bf 0.476 & 0.096 & \bf 0.221 & \bf 0.215 & 0.160 \\
\emph{SILR} & \bf 0.296 & 0.471 & \bf 0.302 & \bf 0.249 & 0.358 & 0.484 & 0.113 & \bf 0.222 & \bf 0.283 & 0.181 \\
\midrule
\multicolumn{11}{c}{ FB15k } \\
\midrule
\emph{GQE} & 0.386 & 0.636 & 0.345 & 0.248 & 0.515 & 0.624 & 0.151 & 0.310 & 0.376 & 0.273 \\
\emph{Q2B} & 0.484 & 0.786 & 0.413 & 0.303 & 0.593 & 0.712 & 0.211 & 0.397 & 0.608 & 0.330 \\
\emph{Q2B+TEMP} & 0.554 & 0.840 & 0.498 & 0.422 & 0.674 & 0.779 & 0.483 & 0.261 & 0.727 & 0.300 \\
\emph{BiQE} & 0.482 & 0.810 & 0.526 & 0.461 & 0.677 & 0.781 & 0.209 & 0.456 & 0.185 & 0.232 \\
\emph{SILR} & \bf 0.596 & \bf 0.867 & \bf 0.575 & \bf 0.489 & \bf 0.736 & \bf 0.824 & 0.287 & \bf 0.500 & \bf 0.767 & 0.321 \\
\midrule
\multicolumn{11}{c}{ NELL995 } \\
\midrule
\emph{GQE} & 0.247 & 0.418 & 0.228 & 0.205 & 0.316 & 0.447 & 0.081 & 0.186 & 0.199 & 0.139 \\
\emph{Q2B} & 0.306 & 0.555 & 0.266 & 0.233 & 0.343 & 0.480 & 0.132 & 0.212 & 0.369 & 0.163 \\
\emph{Q2B+TEMP} & 0.373 & 0.625 & 0.343 & 0.342 & 0.410 & 0.552 & 0.209 & 0.141 & 0.477 & 0.262  \\
\emph{BiQE} & 0.309 & 0.632 & 0.310 & 0.332 & 0.370 & 0.525 & 0.091 & 0.164 & 0.173 & 0.184 \\
\emph{SILR} & 0.342 & \bf 0.641 & 0.329 & 0.337 & 0.376 & 0.532 & 0.055 & 0.177 & 0.450 & 0.182 \\
\bottomrule
\end{tabular}
}
\caption{\label{table_result_transductive} Transductive H@3 results on FB15k-237, FB15k and NELL995 datasets. The results of \emph{GQE} and \emph{Q2B} are taken from~\cite{ren2020query2box}.}
\end{center}
\end{table*}

\subsection{Inductive Setting}
\paragraph{Unseen Entities Generalization} To illustrate the inductive performance of complex logical reasoning over KGs, we first make a comparison on FB15k-237-V2 and NELL-V3 datasets for generalizing to unseen entities. We compare our \underline{\bf s}tructure-modeled \underline{\bf i}nductive \underline{\bf l}ogical \underline{\bf r}easoning method (\emph{SILR}), with the baseline embedding-based method \emph{Q2B}~\cite{ren2020query2box} and the best version of the inductive model \emph{TEMP(GQE)}~\cite{hu2022type}. \emph{TEMP(GQE)} enriches embedding method \emph{GQE}~\cite{hamilton2018embedding} with type information which has achieved the state-of-the-art performance. 
\emph{BiQE}~\cite{kotnis2021answering} is also a textual encoding method with positional embedding for logical query but only in the transductive setting. We reimplement it by replacing the original classification-based prediction with the matching scheme for an inductive comparison.

The experimental results are shown in Table~\ref{table_result_inductive1}. We can see that our \emph{SILR} outperforms all other models on both FB15k-237-V2 and NELL-V3 datasets by a considerable margin. This highlights the effectiveness of our method for inductive logical reasoning over unseen entities. Additionally, the improvement over the other positional textual encoding model, \emph{BiQE}, demonstrates that our structure knowledge modeling during textual encoding is capable of enhancing the inductive complex query answering capability.    
% We also compare with our baseline model without structural prompt and different operation modeling.

\paragraph{Cross-KG Generalization}
We further explore a more challenging cross-KG inductive setting, where the model is trained and tested on different datasets and requires generalizing to completely different KGs. Specifically, we take FB15k and NELL995 as the source/target and target/source datasets, respectively. 
In this scenario, we adopt the few-shot setting, where 500 random samples of the target domain are provided for continual learning to achieve better transferring.
As embedding-based methods, even when aware of type information, are unable to embed most entities and relations in new KGs with unseen types, we only compare our \emph{SILR} with the reimplemented \emph{BiQE}. The results in Table~\ref{table_result_inductive2} show that our \emph{SILR} performs better than \emph{BiQE}, and it can not only generalize to unseen entities but also perform logical reasoning over new KGs with only a few portions observed. 
This manifests the effectiveness of our method on inductive logical reasoning over KGs even in the real-world challenging cross-KG setting.
% 补充是否有few-shot samples training.

\subsection{Transductive Setting} 
Although textual encoding methods have the 
inductive potential, their performance often lags behind embedding-based models due to learning inefficiency and the inability to structure knowledge modeling~\cite{wang2022simkgc}.
We also compare our \emph{SILR} with transductive logical reasoning methods to illustrate the logical reasoning performance over KGs with structure modeling. 
The compared models including \emph{GQE}, \emph{Q2B}, \emph{TEMP(Q2B)} and \emph{BiQE} where the first three are embedding-based models while the last one is a positional textual encoding model. Since \emph{BiQE} does not evaluate query types 2u and up, and does not provide results for FB15k, we reimplement it for a fair comparison.

The results are shown in Table~\ref{table_result_transductive}. 
Our \emph{SILR} outperforms \emph{BiQE}, particularly in query types involving intersection and union, indicating that our structure knowledge modeling can effectively improve the logical reasoning performance of textual encoding and help generalize to unseen query structures.
Although textual encoding methods have the potential for inductive KG reasoning, they still lag behind embedding-based methods for the transductive setting, due to explicit structure modeling and better learning efficiency of embedding-based methods~\cite{wang2022simkgc}.
In this work, we mainly focus on improving textual encoding methods for inductive complex reasoning, but our method still achieves comparable transductive performance. 
% Compared to embedding-based methods with more explicit structure modeling and better learning efficiency, our method still achieves comparable performance. 
This demonstrates the effectiveness of our inductive method with query structure modeling on transductive logical reasoning over KGs.

\subsection{Further Analysis}
\paragraph{Ablation Study}
To dive into the impact of different components in our model on both inductive and transductive logical reasoning over KGs, we conduct an ablation study on the FB15k-237-V2 and FB15k-237 datasets. 
We respectively take bert-large-cased and bert-base-cased as baseline models for FB15k-237-V2 and FB15k-237. They remove Structural Prompt (\emph{SP}) and Geometric Operation Modeling (\emph{GOM}) from the final model \emph{SILR}, which directly encodes linearized query structures for answer prediction.

As shown in Table~\ref{table_ablation}, incorporating structural prompting and geometric operation modeling can both improve the baseline model, but still perform worse than our final \emph{SILR}.
This indicates that these two measures for modeling structure knowledge during query text encoding can enhance the inductive and transductive performance of logical reasoning over KGs.
\begin{table}[!th]
    \begin{center}
    \resizebox{0.48\textwidth}{!}{
    \begin{tabular}{l|cc|cc}
    \toprule 
    \multirow{2}{*}[-0.7ex]{\centering Model}
    &\multicolumn{2}{c}{FB15k-237-V2} 
    &\multicolumn{2}{c}{FB15k-237} \\
    \cmidrule(lr){2-3} \cmidrule(lr){4-5} 
    & H@3 & H@10 & H@3 & H@10 \\
    \midrule
    \emph{Baseline}  & 0.072 & 0.151 & 0.258 & 0.397 \\
    \ \ \ \ \emph{w/ SP} & 0.073 & 0.162 & 0.288 & 0.429 \\
    \ \ \ \ \emph{w/ GOM} & 0.071 & 0.167 & 0.281 & 0.420 \\
    \emph{SILR} & 0.079 & 0.178 & 0.296 & 0.437 \\
    \bottomrule
    \end{tabular}
    % }
    }
    \caption{\label{table_ablation} Ablation study results on FB15k-237-V2 and FB15k-237. \emph{SP} and \emph{GOM} respectively denote Structural Prompt and Geometric Operation Modeling.}
    \end{center}
\end{table}

\paragraph{Query Structure Generalization}
Embedding-based methods are known to generalize well to unseen query structures due to their explicit spatial structure learning. To analyze the generalizability of our implicit structure-modeled textual encoding method to different logical query structures, we further construct four types of complicated query structures with more relations and geometric operations, including 4p, 5p, 3ip and i2p, based on the query structures in test sets.
The detailed illustrations and explanations of these query structures are given in Appendix~\ref{four_query_type}. We directly evaluate our method on these more complicated queries in both inductive and transductive datasets FB15k-237-V2 and FB15k-237. The results are listed in Table~\ref{table_query_structure}. We can see that compared to seen and unseen query structures in the original datasets in Table~\ref{table_result_inductive1} and~\ref{table_result_transductive}, our method can also generalize to these complicated structures with more logical operations and achieves impressive performance. This demonstrates that the design of structural prompt and implicit geometric operation modeling can effectively learn structure knowledge and improve query structure generalizability.

\begin{table}[!th]
	\centering
    \begin{center}
    \resizebox{0.48\textwidth}{!}{
    \setlength{\tabcolsep}{2mm}{
    \begin{tabular}{l|ccccc}
    \toprule 
    Model & 4p & 5p & 3ip & i2p \\
    \midrule
    Inductive & 0.186 & 0.089 & 0.256 & 0.092 \\
    \midrule
    Transductive & 0.192 & 0.188 & 0.367 & 0.199 \\
    \bottomrule
    \end{tabular}
    }
    }
\caption{\label{table_query_structure} Analysis of generalization to  more complicated query structures.
    For inductive, we report the H@10 scores on FB15k-237-V2 and for transductive we report the H@3 results on FB15k-237.}
    \end{center}
\end{table}

% \paragraph{Text Question Answering}

\section{Related Work}
Answering first-order logic queries over incomplete KGs is a challenging task~\cite{guu2015traversing}. Previous research~\cite{lin2015modeling, hamilton2018embedding} mainly studies transductive logical reasoning, where the training and testing are performed on the same KG.
Embedding-based methods are widely used to embed both logical queries and entities into a joint low-dimensional vector space and push answer entities and queries to be close enough, enabling answer prediction through embedding similarity, even when the involved relation is absent. Following this paradigm, some further propose extending the embedding of the query/entity from a single point to a region~\cite{ren2020query2box, zhang2021cone, choudhary2021self} or probabilistic distribution~\cite{ren2020beta,choudhary2021probabilistic} over vector space to map arbitrary first-order logic queries into sets of answer entities. However, these methods are unable to tackle the inductive  problem, which requires generalizing to unseen entities and relations. Although~\cite{hu2022type} proposes enriching the entity and relation embedding with type information for inductive logical reasoning, it can only generalize to elements of observed types. 

Another line of research focuses on inductive logical reasoning over KGs using textual encoding methods. With the advance of large-scale pre-trained language models~\cite{devlin2018bert,liu2019roberta}, these methods propose transforming the graph structures into linearized text and utilize PLMs for encoding~\cite{yao2019kg, zha2022inductive}. With the strong generalizability of PLMs, they can easily generalize to unseen entities/relations, but struggle to model structure knowledge during text encoding. Some works~\cite{wang2021kepler, daza2021inductive, wang2021structure} propose to follow TransE~\cite{bordes2013translating} to apply the translation function between entity and relation representations for geometric structure learning. Nevertheless, 
these methods usually require descriptions of entities and relations for encoding and assume these descriptions are readily available. Besides, 
they only focus on simple link prediction tasks without exploring complex structure modeling in logical reasoning, which is essential for generalizing within the same query structure/type. We thus propose to simultaneously encode the linearized query and preserve the logical structure knowledge by structural prompt and separate geometric operation modeling for inductive logical reasoning over KGs.

\section{Conclusion}
% In this paper, we focus on the task of inductive logical reasoning over KGs, which requires generalizing to unseen elements. 
In this paper, we present the first flexible inductive method for answering complex logical queries over incomplete KGs which can generalize to any unseen entities and relations.
To accomplish this goal, we propose a structure-model textual encoding model that utilizes PLMs to encode linearized query structures to find the answer entities. For structure modeling of complex queries, we design structural prompts to implicitly indicate PLMs the order of geometric operations execution in each query, and separately model three geometric operations on representation space using a pre-trained encoder, an attention layer, and a maxout layer. Experimental results demonstrate the effectiveness
of our model on logical reasoning over KGs in both inductive and transductive settings.

\section*{Limitations}
This study has potential limitations. First, it only focuses on answering existential positive first-order logic queries but does not support the negation operation.  We will later address this limitation by modeling the negation operation.
Second, we utilize BERT as the backbone model for inductive generalization due to computing resource limits. We plan to  investigate the use of more powerful pre-trained language models with stronger generalizability in future research to improve inductive logical reasoning over KGs.

\section*{Acknowledgments}
This work is supported by National Natural Science Foundation of China (No. 6217020551) and Science and Technology Commission of Shanghai Municipality Grant (No.21QA1400600).

\bibliography{anthology,custom}
\bibliographystyle{acl_natbib}

\appendix

\section{Structure Linearization Templates}
\label{templates}
In this part, we list the detailed linearization templates of different structural query types in Table~\ref{table_templates}.
\begin{table*}[!t]
    \begin{center}
    \resizebox{0.94\textwidth}{!}{
    \begin{tabular}{m{2.8cm}m{15cm}}
    \toprule 
    \bf Query Type & \bf Linearization Template \\
    \midrule
    \qquad \ \ \text{[1p]} & [projection] [anchor] $t(e_1)$ [relation] $t(r_1)$ \\
    \midrule
    \qquad \ \ \text{[2p]} & [projection] [anchor] $t(e_1)$ [relation] $t(r_1)$ [projection] [relation] $t(r_2)$ \\
    \midrule
    \qquad \ \ \text{[3p]} & [projection] [anchor] $t(e_1)$ [relation] $t(r_1)$ [projection] [relation] $t(r_2)$ [projection] [relation] $t(r_3)$ \\
    \midrule
    \qquad \ \ \text{[2i]} & [intersection] [projection] [anchor] $t(e_1)$ [relation] $t(r_1)$ [projection] [anchor] $t(e_2)$ [relation] $t(r_2)$  \\
    \midrule
    \qquad \ \ \text{[3i]} & [intersection] [projection] [anchor] $t(e_1)$ [relation] $t(r_1)$ [projection] [anchor] $t(e_2)$ [relation] $t(r_2)$ [projection] [anchor] $t(e_3)$ [relation] $t(r_3)$ \\
    \midrule
    \qquad \ \ \text{[pi]} & [intersection] [projection] [anchor] $t(e_1)$ [relation] $t(r_1)$ [projection] [relation] $t(r_2)$ [projection] [anchor] $t(e_2)$ [relation] $t(r_3)$ \\
    \midrule
    \qquad \ \ \text{[ip]} & [intersection] [projection] [anchor] $t(e_1)$ [relation] $t(r_1)$ [projection] [relation] $t(r_3)$ [projection] [anchor] $t(e_2)$ [relation] $t(r_2)$ [projection] [relation] $t(r_3)$ \\
    \midrule
    \qquad \ \ \text{[2u]} & [union] [projection] [anchor] $t(e_1)$ [relation] $t(r_1)$ [projection] [anchor] $t(e_2)$ [relation] $t(r_2)$  \\
    \midrule
    \qquad \ \ \text{[up]} & [union] [projection] [anchor] $t(e_1)$ [relation] $t(r_1)$ [projection] [relation] $t(r_3)$ [projection] [anchor] $t(e_2)$ [relation] $t(r_2)$ [projection] [relation] $t(r_3)$ \\
    \bottomrule
    \end{tabular}
    }
\caption{\label{table_templates} The linearization templates of each query structure.}
    \end{center}
\end{table*}

\begin{table*}[!th]
    \begin{center}
    \resizebox{0.75\textwidth}{!}{
    \setlength{\tabcolsep}{3mm}{
    \begin{tabular}{l|cc|cc|cc}
    \toprule 
    \multirow{2}{*}[-0.7ex]{\centering Dataset}
    &\multicolumn{2}{c}{Training} 
    &\multicolumn{2}{c}{Validation}
    &\multicolumn{2}{c}{Test}\\
    \cmidrule(lr){2-3} \cmidrule(lr){4-5}  \cmidrule(lr){6-7}
    & 1p & others & 1p & others & 1p & others \\
    \midrule
    FB15k  & 273,710 & 273,710 & 59,097 & 8,000 & 67,016 & 8,000 \\
    FB15k-237 & 149,689 & 149,689 & 20,101 & 5,000 & 22,812 & 5,000 \\
    NELL995 & 107,982 & 107,982 & 16,927 & 4,000 & 17,034 & 4,000 \\
    \midrule
    FB15k-237-V2  & 9,964 & 9,964 & 1,738 & 2,000 & 791 & 1,000 \\
    NELL-V3 & 12,010 & 12,010 & 2,197 & 2,000 & 1,167 & 1,500 \\
    \bottomrule
    \end{tabular}
    }
    }
    \caption{\label{table_statistics} Statistics of logical queries in different types in each data split. ``others'' means other eight query types including 2p, 3p, 2i, 3i, pi, ip, 2u and up.}
    \end{center}
\end{table*}
\section{Data Statistics}
\label{data_statistics}
In Table~\ref{table_statistics}, we summarize the statistics of logical queries in our experimented datasets for both inductive and transductive settings.

% \section{Structure Prompts}
% \label{structural_prompt}
% As shown in Table~\ref{table_prompts}, we present the structural prompts of each query type to instruct the pre-trained language model to implicitly execute involved logical operations stepwise.   
% \begin{table*}[!t]
%     \begin{center}
%     \resizebox{0.92\textwidth}{!}{
%     \begin{tabular}{m{3cm}m{14.5cm}}
%     \toprule 
%     \bf Query Type & \bf Structural Prompt \\
%     \midrule
%     \qquad \ \ \text{[1p]} & One step: [projection] \\
%     \midrule
%     \qquad \ \ \text{[2p]} & Two steps: [projection], then [projection] \\
%     \midrule
%     \qquad \ \ \text{[3p]} & Three steps: [projection], then [projection], then [projection] \\
%     \midrule
%     \qquad \ \ \text{[2i]} & Three steps: [projection], and [projection], [intersection] \\
%     \midrule
%     \qquad \ \ \text{[3i]} & Four steps: [projection], and [projection], and [projection], [intersection] \\
%     \midrule
%     \qquad \ \ \text{[pi]} & Four steps: [projection], then [projection], and [projection], [intersection] \\
%     \midrule
%     \qquad \ \ \text{[ip]} & Five steps: [projection], then [projection], and [projection], then [projection], [intersection] \\
%     \midrule
%     \qquad \ \ \text{[2u]} & Three steps: [projection], and [projection], [union]  \\
%     \midrule
%     \qquad \ \ \text{[up]} & Five steps: [projection], then [projection], and [projection], then [projection], [union] \\
%     \bottomrule
%     \end{tabular}
%     }
% \caption{\label{table_prompts} The structural prompt of each query type.}
%     \end{center}
% \end{table*}

\section{Constructed Query Types}
\label{four_query_type}
We here introduce our generated more complicated query types involving more relations and logical operations, which are used to illustrate the complicated query structure generalizability.
\begin{align}
    [4p] \ q=v_?&.\exists v_1,v_2,v_3:r_1(e_1,v_1)\wedge r_2(v_1,v_2) \nonumber \\ 
    &\wedge r_3(v_2,v_3)\wedge r_4(v_3,v_?) \nonumber \\
    [5p] \ q=v_?&.\exists v_1,v_2,v_3,v_4:r_1(e_1,v_1)\wedge r_2(v_1,v_2) \nonumber \\ 
    &\wedge r_3(v_2,v_3)\wedge r_4(v_3,v_4)\wedge r_5(v_4,v_?) \nonumber \\
    [3ip] \ q=v_?&.\exists v:r_1(e_1,v)\wedge r_2(e_2,v)\wedge r_3(e_3,v) \nonumber \\
    &\wedge r_4(v,v_?) \nonumber \\ 
    [i2p] \ q=v_?&.\exists v_1,v_2:r_1(e_1,v_1)\wedge r_2(e_2,v_1) \nonumber \\
    &\wedge r_3(v_1,v_2)\wedge r_4(v_2,v_?) \nonumber \\ 
\end{align}

\end{document}